\documentclass{article} 
\usepackage{iclr2027_conference,times}

\usepackage{amsmath,amsfonts,bm}

\def\eqref#1{equation~\ref{#1}}

\def\1{\bm{1}}

\DeclareMathAlphabet{\mathsfit}{\encodingdefault}{\sfdefault}{m}{sl}
\SetMathAlphabet{\mathsfit}{bold}{\encodingdefault}{\sfdefault}{bx}{n}

\usepackage{hyperref}
\usepackage{url}

\usepackage{amsmath,amssymb,amsfonts}
\usepackage{booktabs}
\usepackage{graphicx}
\usepackage{xcolor}
\usepackage{multirow}
\usepackage{textcomp}
\usepackage{colortbl}
\usepackage{wrapfig}
\usepackage[normalem]{ulem}
\usepackage{tikz}
\usetikzlibrary{arrows.meta,positioning}

\definecolor{groupbg}{RGB}{245,247,249}
\definecolor{oursrow}{RGB}{232,241,250}
\definecolor{oraclerow}{RGB}{244,248,252}
\newcommand{\up}{\textsuperscript{\color{gray}$\uparrow$}}
\newcommand{\dn}{\textsuperscript{\color{gray}$\downarrow$}}
\newcommand{\ours}{\rowcolor{oursrow}}
\newcommand{\oraclebg}{\rowcolor{oraclerow}}
\newcommand{\second}[1]{\underline{#1}}

\newcommand{\method}{MetaAct}
\newcommand{\gear}{GEAR}

\newcommand{\todo}[1]{}

\title{\method: Self-Evolving Executable Action Constraints for Embodied Language Agents}

\author{
{\bfseries Jiaxing Zhao$^{1}$, Hanwen Wang$^{1}$, Di Liu$^{2}$, Shaoteng Liu$^{3,\dagger}$, Jing-Ran Su$^{4}$, Yanheng Li$^{5}$,}\\
{\bfseries Qian Zha$^{1}$, Jiaxi Li$^{1}$, Hongbin Xie$^{6}$, Haoran Zhang$^{6,*}$, Xuan Song$^{1,*}$, Haoxuan Che$^{7,*}$}\\[0.6em]
{\small $^{1}$Jilin University \quad
$^{2}$Southern University of Science and Technology \quad
$^{3}$Northwestern Polytechnical University}\\
{\small $^{4}$The Hong Kong Polytechnic University \quad
$^{5}$City University of Hong Kong \quad
$^{6}$Peking University}\\
{\small $^{7}$XGEN Labs}\\[0.5em]
{\footnotesize \texttt{\{zhaojx9921,chaqian9923,jxli9925\}@mails.jlu.edu.cn}
\quad \texttt{wanghanwen@jlu.edu.cn}}\\
{\footnotesize \texttt{songxuan@jlu.edu.cn}
\quad \texttt{seucliudi@gmail.com}
\quad \texttt{liushaoteng@live.com}}\\
{\footnotesize \texttt{jing-ran.su@connect.polyu.hk}
\quad \texttt{yanhengli3-c@my.city.edu.hk}}\\
{\footnotesize \texttt{12131108@mail.sustech.edu.cn}
\quad \texttt{h.zhang@pku.edu.cn}
\quad \texttt{hche@xgenlabs.ai}}
}

\iclrfinalcopy 

\begin{document}

\maketitle

\ificlrfinal
\begingroup
\renewcommand{\thefootnote}{\fnsymbol{footnote}}
\footnotetext[1]{Corresponding authors}
\footnotetext[2]{Project leader}
\endgroup
\fi

\begin{abstract}
Self-evolving language agents have emerged as a promising paradigm for continuously improving agent behavior through experience. However, most existing approaches evolve textual artifacts, which can influence action preferences but cannot reliably enforce action admissibility in embodied environments with discrete and often irreversible actions. We propose \method{}, which shifts self-evolution from textual guidance to executable action constraints by synthesizing \gear{} (Grounded Executable Action Rules) from execution feedback. Each rule is represented as a typed specification, compiled into a deterministic guard, and validated offline through trajectory replay without additional model calls or environment interactions. Across four benchmarks, \method{} improves task performance; on ScienceWorld and WebShop, violations drop from 70\% to 0\% and 40\% to 1\%, with score gains of +10.1 and +8.0. Our code is available at \url{https://github.com/BaiCubed/MetaAct}.
\end{abstract}

\section{Introduction}

Language agents can improve by modifying textual artifacts, including specialist roles and
communication structures~\citep{hong2024metagpt,wu2023autogen}, natural-language reflections and
distilled lessons from previous failures~\citep{shinn2023reflexion,zhao2024expel}, and evolving
task contexts~\citep{zhang2026ace}. These methods alter the context
from which the model generates its next output and have shown value in tasks where behavior
is mediated primarily through text. Embodied interaction differs in a consequential way. At
each step, an agent selects a discrete action from an environment-defined set; once executed,
the action changes the environment and may be difficult or impossible to reverse.

\begin{figure}[t]
\centering
\includegraphics[width=\textwidth]{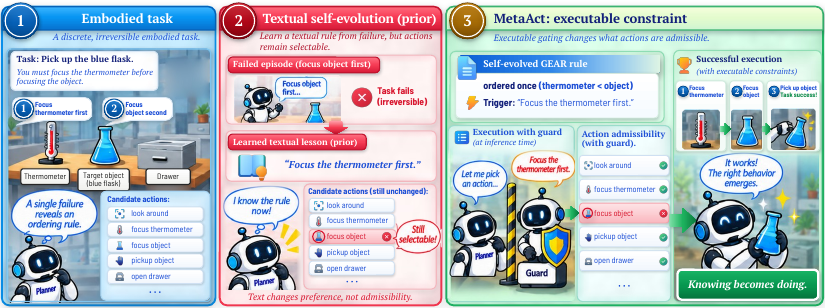}
\caption{\textbf{From textual lessons to executable constraints.} (1)~In ScienceWorld, a single irreversible failure reveals an ordering rule: the thermometer must be focused before the target object. (2)~Textual self-evolution distills the failure into a correct lesson, but the violating action remains in the candidate set and is still selectable. (3)~\method{} compiles the same rule into a \gear{} guard that removes the violating action from the admissible set at execution time: text changes preference, whereas the constraint changes admissibility.}
\label{fig:teaser}
\end{figure}

Figure~\ref{fig:teaser} illustrates this distinction in ScienceWorld, where focusing on the object
under test before the thermometer can terminate the episode immediately. Distilling this failure
into a natural-language lesson gives the planner the correct ordering rule, but leaves the violating
action available. Compiling the same rule into an executable constraint instead removes that action
from the admissible set until its prerequisite is satisfied. Prompt adaptation therefore reshapes
which action the model prefers, whereas executable constraints directly restrict which actions can
be executed, in the spirit of action masking and shielding in reinforcement
learning~\citep{huang2022actionmasking,alshiekh2018shielding}.

To address this gap, we introduce \method{}, a framework that synthesizes executable action
constraints from execution feedback. Rather than letting the model write executable code directly~\citep{wang2023voyager}, each
constraint is represented as a typed \gear{} (Grounded Executable Action Rules) specification and
compiled into a deterministic action guard. Candidate guards are validated offline using previously logged trajectories and, once
accepted, can be reused across episodes without additional environment interactions or LLM
token cost at inference.

We isolate the effect of representation through a controlled comparison in which the failure
evidence, evaluation instances, language model, action budget, and online lesson-evolution
mechanism are held fixed; the only difference is whether the additional evidence is retained
as natural-language guidance~\citep{shinn2023reflexion,automanual,trajbootstrap} or compiled
into an executable constraint. Although the lessons state the correct rules, the episode-level
violation rate is unchanged at $70\%$ on ScienceWorld and $39\%$ versus $40\%$ on WebShop,
whereas the compiled constraints reduce it to $0\%$ and $1\%$ and improve task scores by
$+10.1$ and $+8.0$. Our contributions are:

\begin{enumerate}
    \item We introduce \method{}, a framework that synthesizes typed, executable action
    constraints from execution feedback and compiles them into deterministic guards.    
    \item We develop an offline replay procedure that validates and refines candidate constraints
    from logged trajectories without additional model calls or environment interactions.
   \item We evaluate MetaAct on four embodied-language benchmarks, showing consistent task-level
improvements. On ScienceWorld and WebShop, executable constraints reduce violation rates from
$70\%$ to $0\%$ and from $40\%$ to $1\%$, while improving task scores by $+10.1$ and $+8.0$,
respectively.
\end{enumerate}

\section{Related Work}
\label{sec:related}

\subsection{Textual and Parameter-Space Adaptation}
Most language-agent adaptation changes model context or parameters. Reflections, distilled lessons, and
test-time memories reuse prior experience~\citep{shinn2023reflexion,zhao2024expel,zhang2026ace};
role, topology, and prompt optimization restructure how the model is
queried~\citep{hong2024metagpt,wu2023autogen,metagen,khattab2024dspy}; replanning intervenes during
execution~\citep{sun2023adaplanner,reflact}; and fine-tuning updates weights~\citep{zeng2024agenttuning}.
However, all of these act only through the model's subsequent interpretation of the artifact;
\method{}'s advantage is that feedback is compiled into a deterministic action gate whose effect does
not depend on the model's compliance.

\subsection{Executable Artifacts}
A complementary line generates artifacts that act directly: verified skill libraries~\citep{wang2023voyager},
policy code and situated programs for embodied planners~\citep{liang2023codeaspolicies,singh2023progprompt,liu2023llmp},
affordance-grounded closed-loop feedback~\citep{huang2022innermonologue}, and program libraries validated
by execution~\citep{ellis2021dreamcoder}. However, these methods add executable behavior that must be run to be
assessed; \method{}'s advantage is a subtractive guard whose effect on logged trajectories can be
checked by replay before deployment.

\subsection{Gating and Runtime Enforcement}
Action-space control spans symbolic preconditions~\citep{fikes1971strips}, shielding and action
masking in reinforcement learning~\citep{alshiekh2018shielding,huang2022actionmasking}, and constraints
induced from traces~\citep{yang2007arms,guan2023worldmodel,mordoch2024safepddl}. Closest to ours are
runtime guards for LLM agents: AgentSpec enforces LLM-written rules~\citep{wang2026agentspec} and
OntoGuard checks actions against ontologies mined from demonstrations~\citep{liu2026ontoguard}.
However, both fix their rules before deployment without behavioral verification; \method{}'s advantage
is a gate synthesized from the agent's own execution feedback and validated by replay.

\section{Method: Self-Evolving \gear{}}
\label{sec:method}

\subsection{Typed \gear{} Specifications}

\method{} places the LLM at the specification level rather than letting it write executable code: the model fills a typed specification and a fixed compiler produces the executable artifact. This trades some expressiveness for deterministic compilation and offline validation (Section~\ref{sec:related}).

\begin{figure}[t]
\centering
\includegraphics[width=\textwidth]{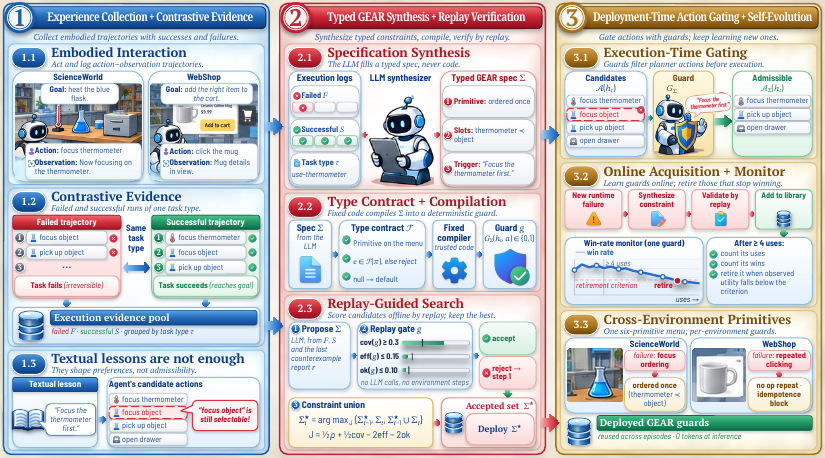}
\caption{\textbf{Overview of \method{}.}
(1)~Failed and successful trajectories provide contrastive execution evidence.
(2)~The LLM synthesizes typed \gear{} specifications, which are compiled into deterministic guards and verified offline by trajectory replay.
(3)~Accepted guards constrain action execution at deployment time and can be continually expanded from new interaction feedback.}
\label{fig:pipeline}
\end{figure}

As illustrated in Figure~\ref{fig:pipeline}, a \gear{} specification contains three components: a rule primitive, typed parameters instantiated through extractor slots, and a trigger message shown to the planner when the rule fires. The primitive is selected from a fixed menu of six operators covering the failure patterns observed during development: \texttt{ordered\_once} enforces a one-time precedence between two action classes (the thermometer ordering in ScienceWorld); \texttt{no\_op\_repeat} blocks an immediate repetition when the observation is unchanged; \texttt{idempotence\_block} rejects an option whose effect has already occurred, including alternating $A$--$B$--$A$--$B$ patterns; \texttt{revisit\_block} forbids returning to a location visited since an anchoring action; \texttt{generic\_deref} resolves a generic reference such as ``the substance'' to the entity implied by the goal; and \texttt{relevance\_filter} restricts an action class to a candidate set extracted from the goal text.

Extractor slots are likewise drawn from a fixed menu of predefined readers over the goal string, so model errors are limited to invalid primitive--extractor combinations and incorrect parameter values, both detectable during compilation and validation. The same restriction bounds coverage: failure patterns outside the primitive menu cannot be constrained (Appendix~\ref{app:coverage}).

\subsection{Executable Action Gating}
\label{sec:gatingview}

A compiled \gear{} specification can be viewed as an execution-time action gate. Let $h_t$ denote
the interaction history, $\mathcal{A}(h_t)$ the actions exposed by the environment, and
$G_{\Sigma}(h_t,a)\in\{0,1\}$ the deterministic decision produced by the compiled specification.
The admissible action set is
\begin{equation}
\mathcal{A}_{\Sigma}(h_t)=\{a\in\mathcal{A}(h_t):G_{\Sigma}(h_t,a)=1\}.
\label{eq:admissible}
\end{equation}
Over a finite candidate set this is equivalent to masking actions with $G_{\Sigma}=0$ before
selection; in free-form action generation, a violating proposal is rejected at the execution interface
and the trigger message is returned for replanning.

Unlike gating in neural systems, the gate here is neither latent nor learned: its implementation and
blocking semantics are fixed, and what evolves is the explicit specification $\Sigma$. \method{} thus
learns \emph{what to gate, not how to gate}. The guarantee is local to the represented constraint:
actions rejected by the current guard cannot pass through it, but failures outside the expressivity or
evidence of $\Sigma$ remain possible.

\subsection{Replay-Based Constraint Verification}

A candidate specification passes through two pre-deployment checks and one post-deployment monitor. The first check is a \textbf{type contract} applied during compilation: a contract $\mathcal{T}$ maps each primitive $\pi$ to the extractor families it admits, $\textsc{compile}$ rejects any specification with $e \notin \mathcal{T}(\pi)$, and \texttt{null} parameters are replaced by declared defaults. This prevents the two implementation failures observed during development, primitive--extractor incompatibility and parameter-type mismatch.

The second check is the \textbf{replay gate}, an empirical verifier in a counterexample-guided inductive synthesis loop~\citep{solarlezama2006sketching}: each candidate is compiled and replayed, and a rejected candidate returns concrete counterexamples to the next synthesis round. Unlike classical CEGIS, no formal specification is assumed; previously logged trajectories serve as the evidence.

Let $F$ and $S$ denote the failed and successful trajectories, respectively, recorded for task type $\tau$ on the training slice. Replaying a compiled guard $g$ over a trajectory produces a blocking decision for each action. We define failure coverage and the effective-action block rate as
\begin{align}
\mathrm{cov}(g) &= \frac{|\{\, f \in F : g \text{ blocks at least one action of } f \,\}|}{|F|},
&
\mathrm{eff}(g) &= \frac{|\{\, a \in A^{+} : g \text{ blocks } a \,\}|}{|A^{+}|},
\label{eq:gate}
\end{align}
where $A^{+}$ is the set of \emph{effective} actions---those that change the observation---in the safety reference
\[
S \cup \{\, f \in F : \mathrm{score}(f) \ge \sigma_{\min} \,\}.
\]
Restricting the denominator of $\mathrm{eff}(g)$ to effective actions prevents no-op actions from being counted as false blocks. Such actions are precisely the behavior that repetition constraints are intended to suppress.

The replay gate accepts a candidate if and only if
\begin{equation}
\underbrace{\mathrm{cov}(g) \ge \tau_{\mathrm{cov}}}_{\text{efficacy}}
\;\wedge\;
\underbrace{\mathrm{eff}(g) \le \tau_{\mathrm{eff}} \;\wedge\; \mathrm{ok}(g) \le \tau_{\mathrm{safe}}}_{\text{safety}},
\label{eq:accept}
\end{equation}
where $\mathrm{ok}(g)$ denotes the block rate over successful trajectories alone. We set $\tau_{\mathrm{cov}}=0.3$, $\tau_{\mathrm{eff}}=0.15$, and $\tau_{\mathrm{safe}}=0.10$; Appendix~\ref{app:gateinternal} analyzes the contribution of the individual acceptance criteria.

All quantities in Eqs.~\ref{eq:gate} and~\ref{eq:accept} are computed by replaying guards over existing logs, so validation needs no additional LLM calls or environment interactions; this is possible because whether a deterministic guard blocks a logged action can be observed directly, whereas the effect of natural-language guidance can only be measured by rerunning the model. A deployment-time \textbf{win-rate monitor} additionally retires an accepted constraint whose observed utility falls below a threshold after four uses; only 2 of the 23 ScienceWorld constraints reach four uses, so it is inactive in all reported results.

\subsection{Replay-Guided Search in Specification Space}

Constraint synthesis is formulated as an iterative search over the specification space, with candidate quality evaluated by the replay gate. At round $t$, the model proposes
\begin{equation}
\Sigma_t \sim \mathrm{LLM}(\tau, F, S, \mathcal{M}, r_{t-1}),
\end{equation}
where $\mathcal{M}$ denotes the available primitive and extractor menu, and $r_{t-1}$ is the counterexample report returned by the previous round. The search uses three complementary mechanisms: contrastive evidence, counterexample-driven repair, and constraint union.

\paragraph{Contrastive evidence.}
The model receives both failed and successful trajectories, allowing it to identify rules that separate ineffective behavior from effective behavior rather than merely describing patterns observed in failures. In an initial WebShop experiment that supplied only failed trajectories, the synthesized candidate blocked both unproductive and successful behavior, mis-blocking $61.4\%$ of successful trajectories. Including successful trajectories provides the evidence needed to distinguish wasteful actions from productive ones.

\paragraph{Counterexample-driven repair.}
When the replay gate rejects a candidate, it produces a report $r_t$ containing concrete examples of incorrectly blocked effective actions and missed wasted steps. These examples are provided directly to the model in the next synthesis round rather than being reduced to a scalar score; unlike text-based optimizers~\citep{yang2024opro,yuksekgonul2024textgrad,agrawal2025gepa}, the signal is generated from the behavior of the compiled artifact on logged trajectories.

\paragraph{Constraint union.}
A single candidate may capture only part of the observed failure pattern. At each round, we therefore evaluate the incumbent specification, the newly proposed specification, and their union:
\begin{equation}
\Sigma^{\star}_t =
\arg\max_{\Sigma \,\in\, \{\Sigma^{\star}_{t-1},\, \Sigma_t,\, \Sigma^{\star}_{t-1} \cup \Sigma_t\}}
J(\Sigma),
\qquad
J = \tfrac{1}{2}\rho + \tfrac{1}{2}\mathrm{cov} - 2\,\mathrm{eff} - 2\,\mathrm{ok},
\label{eq:union}
\end{equation}
where $\rho$ is the fraction of wasted steps detected by the compiled guard. Because the union of two specifications is compiled into the conjunction of their guards, its coverage and false-block rates remain directly measurable through offline replay.

Union lets complementary candidates accumulate across rounds: on WebShop it expands a minimal gate-passing candidate into the final constraint set, and repeated accumulation can grow a specification to 17 constraints for \texttt{use-thermometer} (Figure~\ref{fig:primheat}).

Search terminates when the replay gate accepts the current specification and $\rho \ge 0.8$, or after a maximum of $T=3$ rounds. Appendix~\ref{app:coverage} shows that the third round never produces the first accepted candidate, indicating that successful synthesis typically occurs within the first two rounds.

\section{Experiments}

Our experiments address five questions in turn: (1)~Does \method{} improve performance across
embodied-language benchmarks (Section~\ref{sec:mainresults})? (2)~Why do executable constraints
outperform textual guidance (Section~\ref{sec:why})? (3)~Can synthesized constraints be reliably
verified before deployment (Section~\ref{sec:reliability})? (4)~Does the benefit generalize across
language-model backbones (Section~\ref{sec:crossbackbone})? (5)~Can executable constraints be
acquired and deployed online (Section~\ref{sec:online})?

\subsection{Setup}

\paragraph{Benchmarks.}
We evaluate four text-based embodied environments: ALFWorld (134 \texttt{valid\_unseen}
episodes of ALFRED household tasks in a TextWorld-style simulator~\citep{shridhar2021alfworld,shridhar2020alfred,cote2018textworld}),
SciWorld (30 task types with two held-out variants each, 60 episodes~\citep{wang2022scienceworld}),
TextCraft (60 crafting episodes with success rate and partial-credit score), and WebShop-100k
(100 test instances under a 15-step cap with partial-credit score~\citep{yao2022webshop}).

\paragraph{Baselines.}
Baselines in Table~\ref{tab:baselines} span four families. Single-agent methods: Act-only,
ReAct~\citep{yao2023react}, SC-ReAct with $k=3$ rollouts~\citep{wei2022cot,wang2023selfconsistency},
Reflexion~\citep{shinn2023reflexion}, AdaPlanner~\citep{sun2023adaplanner}, and ReflAct~\citep{reflact}. Orchestration frameworks: AutoGen and MetaGPT-SOP~\citep{wu2023autogen,hong2024metagpt}.
Experience-driven adaptation methods retain, update, or retrieve reusable task experience. Runtime
action guards intercept each proposed action: AgentSpec enforces LLM-generated
trigger--check--enforce rules~\citep{wang2026agentspec}, and OntoGuard checks actions against
ontologies mined from demonstrations~\citep{liu2026ontoguard}. All methods share the planner prompt,
step cap, backend, and temperature; deliberate search over reasoning states~\citep{yao2023tot} is
omitted because it multiplies environment interactions per decision under irreversible actions.

\paragraph{Details.}
The main benchmark comparison uses DeepSeek-V4-flash; cross-backbone experiments additionally
evaluate GPT-4.1-mini and Qwen3-8B under the same SciWorld protocol. The replay-gate thresholds
($\tau_{\mathrm{cov}}=0.3$, $\tau_{\mathrm{eff}}=0.15$, $\tau_{\mathrm{safe}}=0.10$) were
selected once without tuning on the test sets; Appendix~\ref{app:gateinternal} analyzes the gate
criteria and Appendix~\ref{app:coverage} the factors that limit synthesis coverage. Our primary
statistic is the paired difference from the hand-written reference with uncertainty over seeds; we
also report the recovery rate $R=(\text{synth}-\text{w/o})/(\text{oracle}-\text{w/o})$ with its
bootstrap interval, and fidelity, the per-action agreement between synthesized and oracle guards.

\begin{table}[t]
\caption{\textbf{Main benchmark comparison.} Performance of published baselines and \method{}
on four embodied-language benchmarks using the DeepSeek-V4-flash backbone.
\method{}-synth denotes automatically synthesized executable constraints and
\method{}-oracle the hand-written reference. Best values are bold; the second-best distinct
value in each metric column is underlined.}
\label{tab:baselines}
\centering
\small
\setlength{\tabcolsep}{5pt}
\begin{tabular}{l r rr rr rr}
\toprule
& \multicolumn{1}{c}{ALFWorld (134)} & \multicolumn{2}{c}{SciWorld (60)}
& \multicolumn{2}{c}{TextCraft (60)} & \multicolumn{2}{c}{WebShop (100)} \\
\cmidrule(lr){2-2}\cmidrule(lr){3-4}\cmidrule(lr){5-6}\cmidrule(lr){7-8}
& SR\up & SR\up & score\up & SR\up & score\up & SR\up & score\up \\
\midrule
\rowcolor{groupbg}\multicolumn{8}{l}{\textit{Single-agent methods}}\\
Act-only            & 0.799 & 0.167 & 44.3 & 0.617 & 80.6 & 0.180 & 57.6 \\
ReAct               & 0.851 & 0.050 & 27.5 & 0.667 & 72.3 & 0.120 & 21.7 \\
SC-ReAct ($k{=}3$)  & 0.851 & 0.150 & 45.7 & 0.850 & 91.2 & 0.090 & 18.0 \\
Reflexion ($\leq3$) & \textbf{0.955} & 0.217 & 53.5 & 0.750 & 82.7 & 0.210 & 36.2 \\
AdaPlanner          & 0.254 & 0.033 & 28.6 & 0.050 & 16.1 & 0.090 & 16.3 \\
ReflAct             & 0.716 & 0.133 & 44.1 & 0.783 & 83.9 & 0.060 & 9.9 \\
\midrule
\rowcolor{groupbg}\multicolumn{8}{l}{\textit{Agent orchestration frameworks}}\\
AutoGen             & 0.888 & 0.267 & 54.6 & \textbf{0.983} & \textbf{98.3} & 0.100 & 17.4 \\
MetaGPT-SOP         & 0.388 & 0.067 & 38.5 & 0.317 & 38.4 & 0.160 & 41.2 \\
\midrule
\rowcolor{groupbg}\multicolumn{8}{l}{\textit{Experience-driven adaptation}}\\
ACE (ICLR'26)       & 0.925 & 0.117 & 34.2 & 0.667 & 82.4 & 0.170 & 35.3 \\
DC (EACL'26)        & \second{0.948} & 0.183 & 39.1 & 0.617 & 74.5 & 0.180 & 43.6 \\
RPMS (2026)         & 0.940 & 0.167 & 46.9 & 0.750 & 84.6 & \textbf{0.290} & 61.7 \\
SkillRL-SB$^{\S}$   & 0.888 & 0.233 & 52.2 & 0.833 & 89.6 & \textbf{0.290} & 59.0 \\
\midrule
\rowcolor{groupbg}\multicolumn{8}{l}{\textit{Runtime action guards}}\\
AgentSpec (ICSE'26)  & 0.336 & 0.083 & 37.4 & 0.233 & 29.9 & 0.000 & 0.0 \\
OntoGuard (ACL'26)   & 0.299 & 0.067 & 34.1 & 0.783 & 88.7 & 0.040 & 9.3 \\
\midrule
\ours \method{}-synth & \textbf{0.955} & \textbf{0.367} & \textbf{62.1}
                      & \second{0.950} & \second{97.5} & \second{0.270} & \second{62.1} \\
\oraclebg \method{}-oracle & --- & \second{0.300} & \second{55.3} & 0.617 & 76.0 & 0.260 & \textbf{63.6} \\
\bottomrule
\end{tabular}
\end{table}

\subsection{Main Benchmark Results}
\label{sec:mainresults}

Table~\ref{tab:baselines} compares \method{} with representative single-agent methods, agent
orchestration frameworks, experience-driven adaptation methods, and existing runtime action guards
across four embodied-language benchmarks. \method{}-synth achieves the best SciWorld results
($0.367$ success rate and $62.1$ score), matches the best success rate on ALFWorld ($0.955$),
and remains competitive on TextCraft ($0.950$ success rate and $97.5$ score). On WebShop, it
achieves a score of $62.1$, outperforming all published baselines in task score while remaining
close to the best success rate. Notably, \method{} also substantially outperforms existing runtime
guards such as AgentSpec and OntoGuard, indicating that synthesizing and verifying constraints
from execution feedback is more effective than applying fixed pre-deployment rules.

\begin{figure}[t]
\centering
\includegraphics[width=\textwidth]{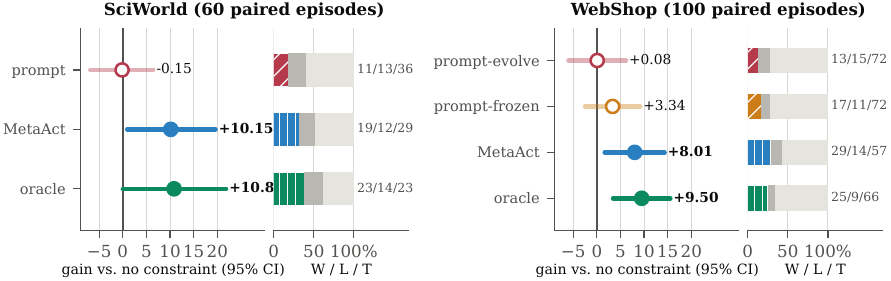}
\caption{\textbf{Matched-evidence comparison.}
Each row shows one variant. Left: task-score change relative to the unconstrained baseline,
with 95\% confidence intervals. Right: episode-level win/loss/tie counts against the same baseline.
Filled markers indicate confidence intervals that do not include zero.}

\label{fig:g1forest}
\end{figure}

\begin{figure}[t]
\centering
\includegraphics[width=\textwidth]{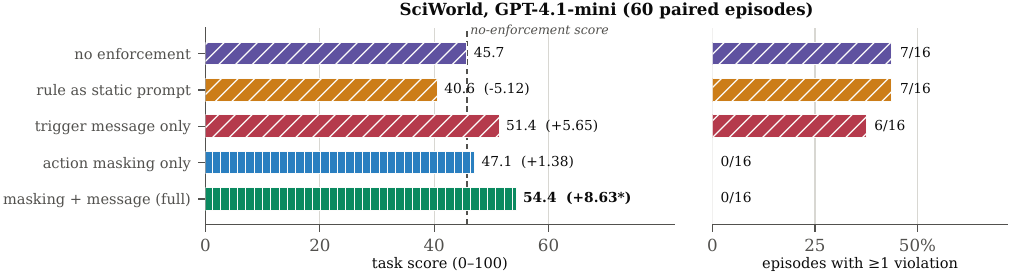}
\caption{\textbf{Effect of action masking and rule messages at execution time.}
Results on SciWorld with GPT-4.1-mini over 60 episodes per setting.
Left: task score under different enforcement strategies ($^*$: the 95\% confidence interval for the score difference is entirely above zero).
Right: fraction of episodes with at least one rule violation.
Action masking removes violations, while combining masking with rule messages gives the largest score improvement.}
\label{fig:mechanism}
\end{figure}

\subsection{Why Do Executable Constraints Outperform Textual Guidance?}
\label{sec:why}

\paragraph{Matched-evidence comparison.}
\label{sec:g1}
Figure~\ref{fig:g1forest} compares textual guidance with executable constraints under matched
evidence. On SciWorld, textual guidance changes the score by only $-0.15$, while MetaAct
improves it by $+10.15$. On WebShop, MetaAct achieves a $+8.01$ gain, whereas neither
prompt-based variant has a confidence interval excluding zero. Across both environments,
executable constraints produce substantially larger gains than representing the same rules as text.

\paragraph{The knowing--doing gap.}
\label{sec:violation}
We further audit the completed trajectories using the same synthesized guard to measure whether
the target rule is actually followed. Despite having the relevant rule explicitly stated in the prompt,
the violation rate remains nearly unchanged: $0.390$ versus $0.400$ on WebShop, and $0.700$
for both the prompt and unconstrained variants on SciWorld. Compiling the same rule into an
executable constraint reduces the violation rate to $0.010$ on WebShop and to $0$ on SciWorld.
This exposes the knowing--doing gap: stating a rule in context does not ensure that the agent
follows it during action selection.

\paragraph{From rule knowledge to rule compliance.}
\label{sec:mechanism}
Figure~\ref{fig:mechanism} examines how the same rule affects behavior under different forms of
enforcement. Providing the rule only as a static prompt leaves violations largely unchanged, while
returning a trigger message improves task performance by $+5.65$ but does not fully prevent
violations. Action masking, in contrast, reduces violations to zero, and combining masking with the
trigger message achieves the largest score gain of $+8.63$. These results highlight the distinction
between knowing a rule and enforcing it: textual feedback can guide the planner, whereas executable
constraints prevent violating actions from being carried out.
Taken together, Figure~\ref{fig:g1forest} establishes the effect, the compliance audit locates its
cause, and Figure~\ref{fig:mechanism} identifies the mechanism: masking guarantees compliance, and the
trigger message helps the planner recover a useful action.

\begin{figure}[t]
\centering
\includegraphics[width=\textwidth]{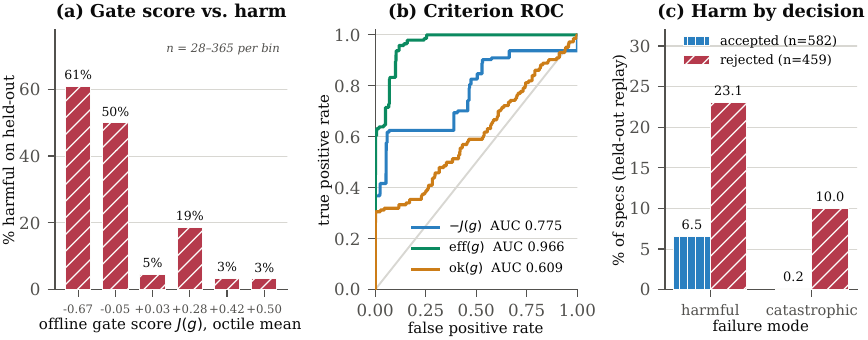}
\caption{\textbf{Offline replay predicts harmful constraint behavior before deployment.}
SciWorld candidate specifications. (a) Held-out harmful blocking as a function of the offline gate
score. (b) ROC curves for the gate criteria. (c) Harmful and catastrophic blocking rates for
accepted and rejected candidates on held-out replay.}
\label{fig:gatecal}
\end{figure}

\begin{table}[t]
\caption{\textbf{Ablation study on SciWorld.}
Offline metrics are computed by replay; online metrics are averaged over three evaluation seeds
with 60 episodes per seed.}
\label{tab:ablmain}
\centering
\small
\begin{tabular}{llrrrrr}
\toprule
& & \multicolumn{3}{c}{Offline} & \multicolumn{2}{c}{Online} \\
\cmidrule(lr){3-5}\cmidrule(lr){6-7}
& Variant & Coverage\up & Fidelity $F_1$\up & Harmful\dn & Violation\dn & Score\up \\
\midrule
\ours A0 & \textbf{Full \method{}} & 0.211 & 0.135 & 0.000 & \textbf{0.167} & \textbf{50.77} \\
A1 & $-$ Contrastive Evidence & 0.189 & 0.411 & 0.000 & 0.229 & 47.62 \\
A2 & $-$ Counterexample Repair & 0.211 & 0.164 & 0.000 & 0.229 & 50.66 \\
A3 & $-$ Constraint Union & 0.189 & 0.164 & 0.000 & 0.292 & 47.37 \\
A4 & $-$ Replay Verification & \textbf{0.433} & 0.135 & 0.069 & 0.229 & 48.36 \\
\bottomrule
\end{tabular}
\end{table}

\subsection{Reliability of Constraint Synthesis and Verification}
\label{sec:reliability}

A synthesized constraint is useful only if a wrong one can be caught before it blocks correct actions.
We therefore ask whether replay verification identifies harmful constraints, and whether each
component of the synthesis pipeline contributes to the deployed result.

\paragraph{Replay verification.}
\label{sec:gatecal}
Figure~\ref{fig:gatecal} evaluates whether offline replay can identify harmful constraints before
deployment. The effective-action block rate $\mathrm{eff}(g)$ is the strongest predictor of held-out
harm, with an AUROC of $0.966$. Accepted constraints show substantially lower harmful blocking
than rejected ones ($6.5\%$ vs. $23.1\%$), while catastrophic blocking drops from $10.0\%$ to
$0.2\%$. These results show that replay verification provides a useful pre-deployment filter for
unsafe constraints.

\paragraph{Ablation analysis.}
\label{sec:ablmain}
Table~\ref{tab:ablmain} evaluates the four components used to build and verify executable
constraints. The full method achieves the lowest violation rate ($0.167$). Removing contrastive
evidence or constraint union increases violations and lowers task performance, showing that both
help produce more effective constraints. Replay verification plays a different role: without it,
coverage increases from $0.211$ to $0.433$, but harmful blocking also appears ($0.069$), indicating
that the additional coverage comes partly from blocking correct actions. Counterexample repair has
a smaller effect, but removing it still increases the violation rate from $0.167$ to $0.229$.

\begin{figure}[t]
\centering
\includegraphics[width=\textwidth]{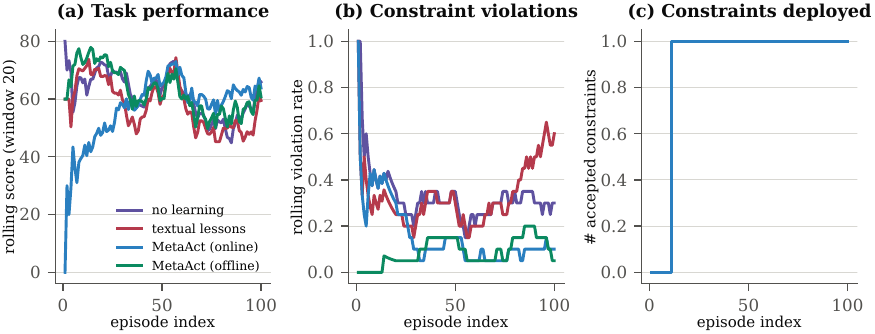}
\caption{\textbf{Executable constraints can be acquired and deployed during an interaction stream.}
WebShop with GPT-4.1-mini over 100 consecutive episodes. The panels show 20-episode rolling task
score, rolling violation rate, and the cumulative number of accepted constraints.}
\label{fig:online}
\end{figure}

\begin{table}[t]
\caption{\textbf{Executable-constraint gains persist across language-model backbones.} SciWorld
task score (60 paired episodes per backbone) with no added constraint, the textual prompt rule,
\method{}, and the hand-written reference.}
\label{tab:crossbackbone}
\centering
\small
\setlength{\tabcolsep}{4.5pt}
\begin{tabular}{l rrrr l}
\toprule
Backbone & No constr. & Prompt rule & \method{} & Oracle & $\Delta$ vs.\ prompt [95\% CI] \\
\midrule
DeepSeek-V4-flash & 44.45 & 44.30 & \textbf{54.60} & 55.30 & $+10.30$ $[+1.15, +19.83]^*$ \\
GPT-4.1-mini      & 45.72 & 40.60 & \textbf{54.35} & 50.83 & $+13.75$ $[+6.43, +21.92]^*$ \\
Qwen3-8B          & 34.77 & 30.90 & \textbf{36.73} & 32.97 & $+5.83$ $[-0.77, +12.93]$ \\
\bottomrule
\end{tabular}
\end{table}

\subsection{Cross-Backbone Generalization}
\label{sec:crossbackbone}

Table~\ref{tab:crossbackbone} evaluates whether the benefit of executable constraints
generalizes across language-model backbones. Under the same SciWorld protocol, we compare
DeepSeek-V4-flash, GPT-4.1-mini, and Qwen3-8B with no added rule, a textual prompt rule,
\method{}, and a hand-written reference. \method{} outperforms the prompt rule on all three
backbones, with confidence intervals excluding zero for DeepSeek-V4-flash and GPT-4.1-mini.
The gain varies across models, but the advantage of executable enforcement is not specific to the
DeepSeek backbone.

\subsection{Online Self-Evolution of Executable Constraints}
\label{sec:online}

Figure~\ref{fig:online} evaluates whether \method{} can acquire executable constraints during
deployment. Starting with no constraint, the agent accumulates failure evidence, synthesizes and
verifies rules, and deploys accepted constraints as the interaction stream proceeds. Online
\method{} reduces the episode violation rate to $0.130$, compared with $0.310$ for
no learning and $0.350$ for textual lesson evolution. Its rolling score also rises from $48.40$
in the first 20 episodes to $63.83$ in the last 20, showing that executable constraints can be
learned and put into use during ongoing interaction.
\method{} therefore not only uses executable constraints; it acquires and deploys them during
ongoing interaction.

\section{Conclusion}

We presented \method{}, a framework that extends language-agent self-improvement from textual
guidance to executable action constraints. \method{} synthesizes typed \gear{} rules from execution
feedback, compiles them into deterministic guards, and validates them through offline trajectory
replay before deployment. Across four embodied-language benchmarks, \method{} consistently
improves task performance; on SciWorld and WebShop, executable constraints reduce target-rule
violations from $70\%$ to $0\%$ and from $40\%$ to $1\%$, while improving task scores by
$+10.1$ and $+8.0$, respectively. These results show that self-evolution can extend beyond textual adaptation: execution feedback can
be converted into reusable constraints that directly shape future action selection.


\bibliographystyle{iclr2027_conference}
\bibliography{iclr2027_conference}

\appendix

\section{Additional Experiments}
\label{app:additional_experiments}

\subsection{State-Grounded Reasoning Baseline}
\label{sec:reflact}

ReflAct~\citep{reflact} tests whether more strongly grounded textual reasoning can address the
same failure mode. Its backbone replaces free-form planning with continuous reflection on the
agent's state relative to the goal. We adopt the backbone and hold the instruction,
few-shot examples, environment interface, step cap, backend, and temperature fixed relative
to our ReAct baseline.

\begin{table}[h]
\caption{ReflAct compared with ReAct under matched settings.}
\label{tab:reflact}
\centering
\small
\begin{tabular}{lrrrr}
\toprule
SciWorld test60 & SR & score & steps & tokens/ep \\
\midrule
ReAct & 0.050 & 27.5 & 24.5 & 43.9k \\
ReflAct & 0.050 & \textbf{31.9} & \textbf{17.9} & 48.7k \\
\midrule
AutoGen (strongest baseline) & 0.267 & 54.6 & 15.0 & 27.7k \\
constraint variant (ours) & \textbf{0.300} & \textbf{55.3} & 33.2 & 139.9k \\
\bottomrule
\end{tabular}
\end{table}

Relative to ReAct, ReflAct improves partial-credit score by $4.4$ points and reaches that score
with 27\% fewer environment steps. The success rate remains $3/60$ in both conditions, and
the ReflAct score remains approximately 23 points below the constrained variant. These results
suggest that better state-grounded reasoning can improve progress without preventing the
irreversible focus error targeted by the action constraint.

ReflAct's published gains use ALFWorld and a different backend, so this comparison should be
interpreted only within the present SciWorld configuration.

\subsection{Robustness to Goal Rewriting}
\label{app:paraphrase}

We replay the same compiled guards while rewriting only the natural-language goal. Four
transformations preserve the task semantics; a fifth deliberately changes the meaning.

\begin{table}[t]
\caption{\textbf{Guard stability under goal rewriting.} SciWorld: 39 episodes, 715 actions, and
63 guard firings under the original goals. The same compiled guards are replayed after each rewrite.}
\label{tab:paraphrase}
\centering
\small
\begin{tabular}{lrrr}
\toprule
Goal perturbation & Consistency\up & Jaccard\up & $\Delta$ blocks \\
\midrule
paraphrase & 1.0000 & 1.000 & +0 \\
clause reordering & 1.0000 & 1.000 & +0 \\
synonym substitution & 1.0000 & 1.000 & +0 \\
distractor clause & 1.0000 & 1.000 & +0 \\
\midrule
meaning-changing control & 0.9692 & 0.774 & +22 \\
\bottomrule
\end{tabular}
\end{table}

The four meaning-preserving rewrites leave every blocking decision unchanged, whereas changing the
goal semantics produces 22 additional blocks. The nontrivial response comes from goal-dependent
primitives such as \texttt{relevance\_filter}; repetition guards that do not read the goal are invariant
by construction. The result therefore shows stability to surface form while preserving sensitivity to
the task semantics used by the guard.

\subsection{Ablation of Replay-Gate Criteria}
\label{app:gateinternal}

To identify which replay statistics carry the selection signal, we re-evaluate the same candidate pool
while changing only the acceptance criterion.

\begin{table}[t]
\caption{\textbf{Contribution of the replay-gate criteria on SciWorld.} The candidate pool is fixed;
only the acceptance rule changes.}
\label{tab:gateinternal}
\centering
\small
\begin{tabular}{lrrrrr}
\toprule
Gate variant & Pass rate & Coverage & False block\dn & Harmful\dn & Catastrophic\dn \\
\midrule
\ours \textbf{Full gate} & 0.643 & 16/23 & 0.0248 & 0.188 & 0.000 \\
$-$ \texttt{cov} & 0.829 & 20/23 & 0.0198 & 0.150 & 0.000 \\
$-$ \texttt{eff} & 0.700 & 19/23 & 0.0208 & 0.316 & 0.000 \\
$-$ \texttt{ok} & 0.643 & 16/23 & 0.0248 & 0.188 & 0.000 \\
$-$ \texttt{eff} \& \texttt{ok} & 0.814 & 20/23 & 0.0231 & 0.300 & 0.000 \\
no gate & 1.000 & 23/23 & 0.0843 & 0.261 & 0.087 \\
\bottomrule
\end{tabular}
\end{table}

The effective-action criterion \texttt{eff} carries the strongest safety signal: removing it raises the
harmful rate from $0.188$ to $0.316$. Removing \texttt{ok} leaves the result unchanged in this
SciWorld pool because successful reference trajectories are sparse. The coverage threshold is
conservative in this setting: removing it admits four additional task types without increasing harm.
The no-gate condition is the only setting that admits catastrophic candidates, with rate $0.087$.

\subsection{Replay-Gate Robustness to Corrupted Evidence}
\label{app:noisy}

We corrupt the validation evidence while keeping the candidate constraints fixed, then evaluate the
accepted set on clean held-out trajectories.

\begin{figure}[t]
\centering
\includegraphics[width=\textwidth]{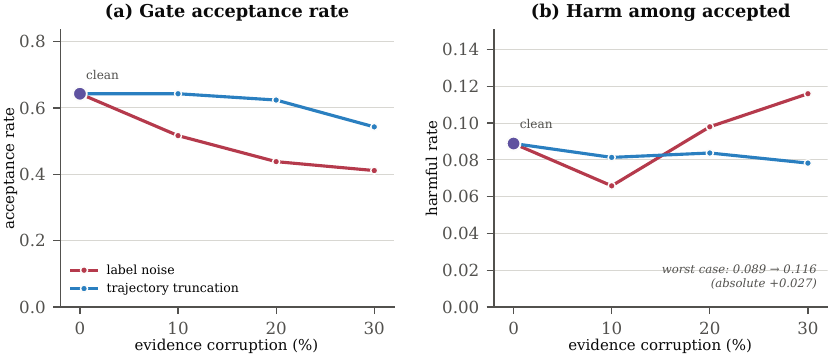}
\caption{\textbf{Replay validation remains selective under corrupted evidence.} SciWorld candidate
constraints under label noise and trajectory truncation. (a) Gate acceptance rate. (b) Harmful rate
among the accepted constraints, evaluated on clean held-out trajectories.}
\label{fig:noisy}
\end{figure}

With 30\% label noise, acceptance falls from $0.643$ to $0.411$, while the harmful rate among
accepted candidates changes from $0.089$ to $0.116$. At 30\% trajectory truncation, acceptance is
$0.543$ and the harmful rate is $0.078$. The dominant response to corrupted validation data is thus
increased selectivity rather than a large increase in held-out harmful blocking.

\subsection{Scaling Behavior under Constraint Accumulation}
\label{app:accumulation}

We stress-test conjunctions of increasingly many verified constraints to separate compiler-level
composability from behavioral over-restriction.

\begin{figure}[t]
\centering
\includegraphics[width=\textwidth]{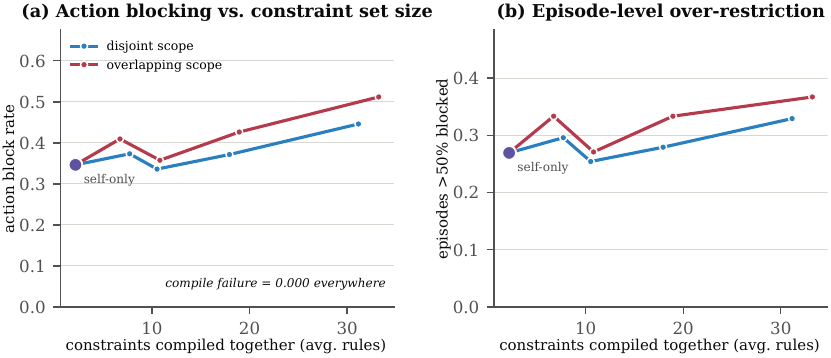}
\caption{\textbf{Constraint accumulation preserves compiler composability while overlapping scopes
increase behavioral restriction.} SciWorld. (a) Action block rate as the number of compiled rules
grows. (b) Fraction of episodes in which more than half of the actions are blocked.}
\label{fig:accumulation}
\end{figure}

Compilation succeeds in every tested setting, including unions of roughly 33 rules. The action block
rate rises from $0.3465$ for the self-only specification to $0.4461$ for large compatible unions and
$0.5120$ when scopes overlap. The corresponding fraction of heavily blocked episodes rises to
$0.367$ in the overlapping condition. The scaling limit observed here is therefore behavioral
scope overlap rather than syntactic or type-level incompatibility.

\subsection{Compliance under Localized Rule Feedback}
\label{app:compliance}

When a guard fires, its message is written to \texttt{last\_error} and inserted verbatim into
the actor and planner prompts on the next turn. The agent then receives two retries. This
condition minimizes context distance and provides the exact rule at the time of violation, so
the fraction of guard activations resolved within the retry budget provides a direct measure
of compliance.

\begin{table}[t]
\caption{Compliance when the exact rule is supplied at the time of violation with two
retries.}
\label{tab:compliance}
\centering
\small
\begin{tabular}{lrrr}
\toprule
Variant & complied within budget & budget exhausted & compliance \\
\midrule
SciWorld, synthesized & 60 & 111 & \textbf{35.1\%} \\
SciWorld, hand-written & 116 & 234 & \textbf{33.1\%} \\
WebShop, synthesized & 52 & 7 & 88.1\% \\
WebShop, hand-written & 64 & 11 & 85.3\% \\
\bottomrule
\end{tabular}
\end{table}

Compliance is 35.1\% for synthesized SciWorld guards and 33.1\% for hand-written SciWorld
guards, compared with 88.1\% and 85.3\% on WebShop. The SciWorld result directly exposes the
knowing--doing gap under a maximally local intervention: even when the exact rule is supplied at the
moment of violation, most attempts are not corrected within the retry budget. This motivates
enforcement at the action interface rather than relying on textual compliance alone.

\section{Main-Result and Efficiency Details}
\label{app:main_details}

\subsection{Three-Seed Comparison with the Hand-Written Reference}
\label{app:three_seed}

\begin{table}[h]
\caption{\textbf{Three-seed comparison with the hand-written reference.} Results use seeds
42/43/44 and are paired by episode ID. $R=(\mathrm{synth}-\mathrm{w/o})/(\mathrm{oracle}-\mathrm{w/o})$
measures recovery of the hand-written headroom.}
\label{tab:main}
\centering
\footnotesize
\setlength{\tabcolsep}{2.5pt}
\begin{tabular}{lrrrlr}
\toprule
& w/o & synth\up & oracle & $R$\up (95\% CI) & Tokens/ep\dn \\
\midrule
SciWorld test60 & 41.8 $\pm$ 3.5 & 55.3 $\pm$ 5.9 & 54.6 $\pm$ 0.9
  & 0.95 $[0.26, 3.78]$ & 101.3k / 137.6k ($-26\%$) \\
WebShop test100 & 55.8 $\pm$ 1.5 & 60.4 $\pm$ 1.5 & 61.4 $\pm$ 2.0
  & 0.78 $[-2.83, 4.75]$ & 19.4k / 21.0k ($-8\%$) \\
\bottomrule
\end{tabular}
\end{table}

The same evaluation instances are repeated across seeds, so the seed is the unit of replication.
The per-seed differences (synth $-$ oracle) are $+6.8/-1.4/-3.4$ on SciWorld and
$-1.5/-1.0/-0.4$ on WebShop, yielding mean differences of $+0.67$ and $-0.97$, respectively.
On WebShop the lower confidence bound remains within the predefined three-point non-inferiority
margin. SciWorld has a wider seed-level interval, so its aggregate similarity to the hand-written
reference is less precisely estimated.

\subsection{Performance Stratified by Synthesis Decision}
\label{app:strata}

\begin{table}[h]
\caption{\textbf{SciWorld performance stratified by whether synthesis emits a verified constraint.}
Strata are determined from training evidence; intervals are computed over the three evaluation seeds.}
\label{tab:strata}
\centering
\small
\begin{tabular}{lrrrr}
\toprule
Stratum & types & per seed & synth $-$ oracle\up & 95\% CI (seeds) \\
\midrule
Constraint emitted & 10 & $-0.8/-5.8/-9.2$ & $-5.27$ & $[-15.8, +5.2]$ \\
Abstained          & 20 & $+10.6/+0.8/-0.4$ & $+3.67$ & $[-11.3, +18.7]$ \\
\midrule
Pooled             & 30 & $+6.8/-1.4/-3.4$ & $+0.67$ & $[-12.8, +14.1]$ \\
\bottomrule
\end{tabular}
\end{table}

The synthesizer emits constraints for 10 of 30 task types and abstains on the remaining 20. On the
emitted types, the hand-written reference has the higher point estimate; on the abstained types, the
unconstrained synthesized pipeline has the higher point estimate. The pooled result therefore combines
partial constraint quality with selective abstention rather than assuming that a rule must be emitted
for every task type.

\subsection{Per-Episode Token Use in the Main Benchmark}
\label{app:tokens}

\begin{table}[h]
\caption{\textbf{Per-episode model-token use for the methods in Table~\ref{tab:baselines}.}}
\label{tab:baseline_tokens}
\centering
\small
\setlength{\tabcolsep}{5pt}
\begin{tabular}{lrrrr}
\toprule
Method & ALFWorld & SciWorld & TextCraft & WebShop \\
\midrule
Act-only & 32.9k & 22.2k & 31.9k & 11.2k \\
ReAct & 40.2k & 43.9k & 49.8k & 32.6k \\
SC-ReAct ($k{=}3$) & 116.0k & 113.7k & 106.9k & 89.6k \\
Reflexion ($\leq3$) & 58.3k & 138.4k & 91.5k & 97.3k \\
AdaPlanner & 4.8k & 4.1k & 4.1k & 39.8k \\
AutoGen & 34.2k & 27.7k & 26.7k & 41.4k \\
MetaGPT-SOP & 19.3k & 15.2k & 21.1k & 39.0k \\
ReflAct & 79.0k & 44.2k & 49.1k & 66.2k \\
ACE (ICLR'26) & 44.6k & 34.9k & 39.4k & 35.5k \\
DC (EACL'26) & 39.2k & 26.0k & 47.8k & 34.1k \\
RPMS (2026) & 45.4k & 62.7k & 48.3k & 16.7k \\
SkillRL-SB$^{\S}$ & 44.6k & 42.2k & 47.4k & 21.4k \\
AgentSpec (ICSE'26) & 34.0k & 63.1k & 5.4k & 28.2k \\
OntoGuard (ACL'26) & 66.2k & 81.8k & 50.6k & 42.4k \\
\method{}-synth & 21.3k & 84.7k & 113.6k & 18.6k \\
\method{}-oracle & --- & 139.9k & 80.8k & 22.8k \\
\bottomrule
\end{tabular}
\end{table}

Token use reflects the full interaction loop rather than the cost of evaluating a guard. Methods with
few model calls can be inexpensive but weak, while constraint-triggered replanning can increase token
use even when the compiled guard itself is deterministic and token-free.

\subsection{Computational Cost}
\label{sec:cost}

Constraint synthesis is a one-time cost per task type, and evaluating a compiled guard itself
requires no LLM call. Total episode-level token use can nevertheless change because blocking an
action may alter the trajectory and trigger additional planning or retries. On SciWorld, synthesis
uses 487k tokens across 23 task types (21k per type); on WebShop it uses 33k tokens for the single
task type. The per-episode token totals for the main baseline comparison are reported in
Appendix~\ref{app:tokens}. Because blocking can trigger corrective replanning or retries, fewer
environment actions do not necessarily imply fewer model tokens; we therefore report token use
separately from task-level performance.

\section{Constraint Synthesis and Guard Analysis}
\label{app:constraint_analysis}

\subsection{Behavioral Stability Across Synthesis Runs}

We distinguish textual reproducibility of the emitted specification from behavioral
reproducibility after compilation. Across three independent synthesis runs with identical
evidence and settings, only 2 of 20 shared task types receive the same primitive set in all
three runs, and 12 of 20 receive the same replay-gate verdict.

Despite this textual variation, the compiled specifications agree on 86.1\% of per-action
decisions ($949/1102$) when replayed over the same 60 held-out trajectories. The replay gate
evaluates candidate behavior rather than exact textual form, so different specifications may
be accepted when they implement similar blocking decisions.

Most residual disagreement is concentrated in
\texttt{lifespan-longest-lived} (49.2\% agreement),
\texttt{inclined-plane-determine-angle} (59.0\%), and
\texttt{mendelian-genetics-unknown-plant} (62.5\%). These are also task types for which
candidates frequently block no recorded failure action and are rejected by the gate
(Appendix~\ref{app:coverage}). Because synthesis is performed once per task type, a deployed
system can cache the accepted specification; run-to-run variation affects reproduction of the
synthesis procedure rather than execution of an already verified artifact.

\subsection{Agreement with Hand-Written Reference Guards}

On WebShop, the synthesized and hand-written guards agree on all 1307 held-out actions,
yielding $P=R=F_1=1.000$ with no false positives or false negatives. On SciWorld, under the
focus-isolated evaluation, precision is $1.000$ and recall is $0.812$. Across three
independent synthesis runs, precision remains $1.000/1.000/1.000$, with constraints covering
an average of 12.3 task types (range 11--13). The synthesized SciWorld guards are therefore
conservative in this evaluation: observed disagreement arises from missed blocks rather than
false blocks.

\subsection{Cross-Environment Primitive Utilization}

The same six-primitive menu is available in every environment, while the selected
specifications are determined separately for each task type.

\begin{figure}[h]
\centering
\includegraphics[width=0.82\textwidth]{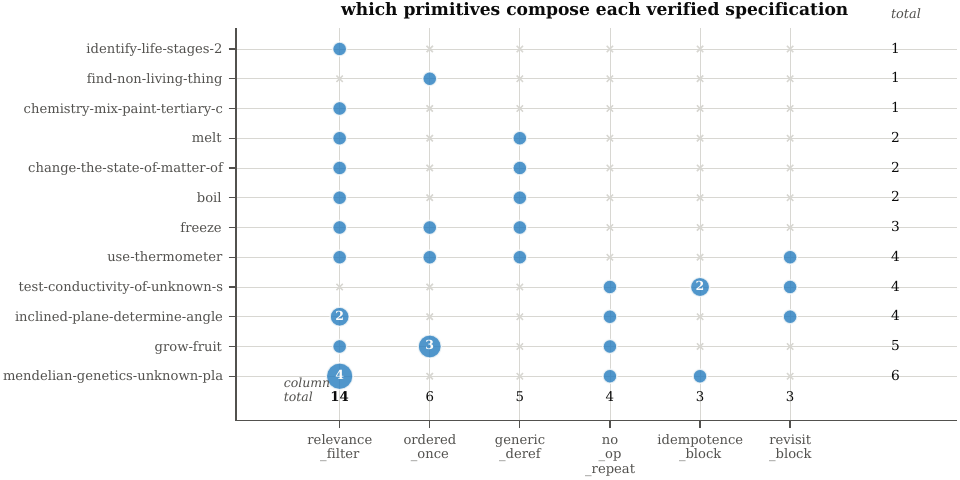}
\caption{Composition of verified specifications. Marker area indicates the number of
constraints using each primitive; $\times$ indicates that a primitive is absent. Column
totals report aggregate use.}
\label{fig:primheat}
\end{figure}

The repetition-related primitives \texttt{no\_op\_repeat} and
\texttt{idempotence\_block} are selected in both SciWorld and WebShop. The remaining
primitives are selected only in environments containing the corresponding ordering,
reference-resolution, or navigation failures. No specification is emitted on ALFWorld,
consistent with its lack of failure evidence under the saturated backbone.

The replay gate also removes cross-environment misapplications. Under the strict gate,
SciWorld selections of \texttt{idempotence\_block} and \texttt{revisit\_block} each decrease
from 3 to 0, while ordering and reference-resolution primitives are retained or become more
frequent. Thus, primitive selection is driven by both the task-level proposal and empirical
validation on the environment's trajectories.

This comparison includes two environments that emit specifications, and WebShop contributes
only one task type; the usage pattern should therefore be interpreted as descriptive rather
than as a broad cross-environment estimate.

\subsection{Analysis of Synthesis Coverage}
\label{app:coverage}

We evaluate three potential explanations for partial coverage: primitive expressiveness,
replay-gate strictness, and the amount of failure evidence.

First, adding a primitive targeted at a common rejected diagnosis does not increase aggregate
coverage. Seven of ten gate-rejected candidates describe the agent leaving the task-relevant
location, a pattern not expressible by the original five-primitive menu. We therefore add
\texttt{revisit\_block}, which keeps the agent at the location associated with the anchoring
action, and rerun synthesis with the primitive menu as the only change. The new primitive is
selected for 2--3 task types per run and corrects specific failures, but the six-primitive
menu verifies $8/9/9$ of the 16 shared types, compared with $8/9/10$ for the five-primitive
menu. The mean difference is $-0.33$ types relative to a within-condition standard deviation
of $0.82$.

Second, relaxing the replay gate provides limited additional coverage. Re-evaluating all
112 specifications generated across five synthesis runs shows that 71\% of rejections occur
because the candidate blocks no action in the failure trajectories. Reducing
$\tau_{\mathrm{cov}}$ from $0.3$ to $0$ therefore admits no additional task type. Removing
both safety criteria increases the specification-level pass rate from $50.9\%$ to $65.2\%$,
but does not resolve the dominant zero-block failure.

Third, the amount of recorded evidence does not separate firing and zero-block candidates.
Both groups contain a median of two failure episodes, with means of $1.82$ and $1.85$,
respectively. The zero-block group also contains slightly more blockable wasted steps
($2.7$ versus $2.2$).

Together, these analyses do not identify primitive vocabulary, gate thresholds, or evidence
count as the primary determinant of coverage in the tested runs. The remaining difficulty is
the mapping from a natural-language diagnosis to a specification that fires on the recorded
failure. For example, \texttt{chemistry-mix} and
\texttt{measure-melting-point-unknown-substance} produce zero-block candidates in all five
runs. Improving diagnosis-to-specification grounding is therefore a direct target for future
work.

\section{Evaluation Protocol and Metric Diagnostics}
\label{app:evaluation_protocol}

\subsection{Dynamics of Online Lesson Memory}
\label{app:lessonmaintenance}

Table~\ref{tab:churn} summarizes how the lesson pool changes during evaluation.

\begin{table}[h]
\caption{Online lesson evolution under the evaluation protocols.}
\label{tab:churn}
\centering
\small
\begin{tabular}{llrrr}
\toprule
Benchmark & Variant & Episodes & Online additions/ep & Median \texttt{uses} \\
\midrule
\multirow{2}{*}{SciWorld} & w/o & 60 & 0.78 & 1 \\
 & prompt (seeded) & 60 & 0.18 & 0 \\
\midrule
\multirow{3}{*}{WebShop} & w/o & 100 & 0.73 & 4 \\
 & prompt-evolve & 100 & 0.76 & 6 \\
 & prompt-frozen & 100 & 0.00 & 100 \\
\bottomrule
\end{tabular}
\end{table}

On WebShop, the retirement rule retains a lesson only when its win rate is at least 0.5 after
three uses, while the base success rate is 0.22. The lesson pool consequently turns over at
0.76 additions per episode, producing a score change of $+0.08$ with 6\% more tokens.
Freezing the same initial lessons yields $+3.3$, indicating that the retirement rule removes
potentially useful lessons together with unhelpful ones.

On SciWorld, each task type has only two test episodes, fewer than the three uses required for
retirement. The seeded pool is already at capacity, so online additions are also limited
(0.18 per episode, with median \texttt{uses} equal to 0). In this setting, the maintenance
mechanism is largely inactive. Accepted executable constraints do not require the same online
retirement loop because they are validated before deployment and then held fixed.

\subsection{Protocol Controls for Retries and Lesson Retirement}
\label{app:protocol}

The unguarded variants also record retries, but those retries result from parsing and grounding
failures rather than action-constraint violations. Their rates (88.6\% for SciWorld w/o and
79.5\% for SciWorld prompt) therefore measure a different event and are not directly
comparable with Table~\ref{tab:compliance}.

The lesson-retirement rule keeps a lesson while \texttt{uses < 3} or when
\texttt{win-rate $\geq 0.5$}. On SciWorld, each task type has only two test episodes, so
retirement never occurs and seeded lessons persist. On WebShop, there is one task type and
the success rate is 0.22, causing each lesson to be removed after reaching three uses unless
it happens to exceed the 0.5 threshold. We observe 76 re-distillations across 100 episodes.
The \texttt{--freeze\_lessons} condition aligns WebShop with SciWorld's effective protocol;
all three initial lessons remain available for all 100 episodes
(\texttt{uses}=100, with no online additions).

\subsection{Violation Timing and Metric Selection}
\label{app:violation}

\begin{figure}[t]
\centering
\includegraphics[width=\textwidth]{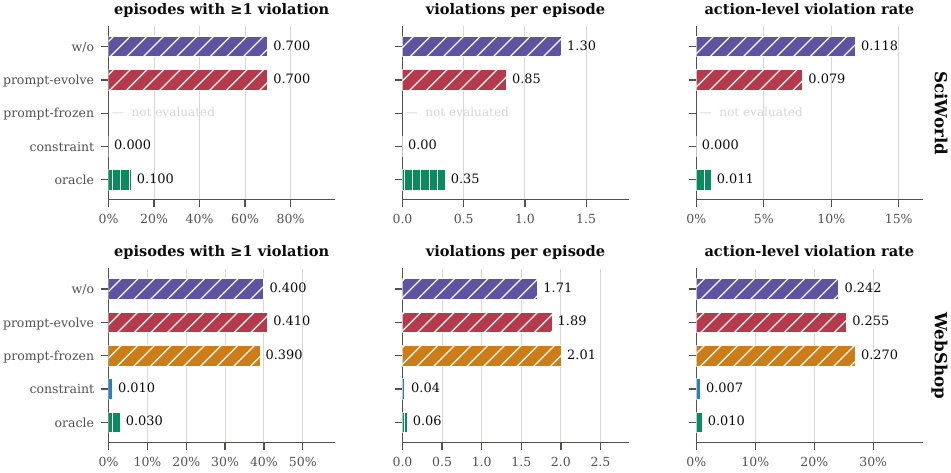}
\caption{Episode-level rule-violation rates. Diagonal hatching denotes rules provided in the
prompt, and grid hatching denotes compiled constraints.}
\label{fig:violation}
\end{figure}

\begin{figure}[t]
\centering
\includegraphics[width=\textwidth]{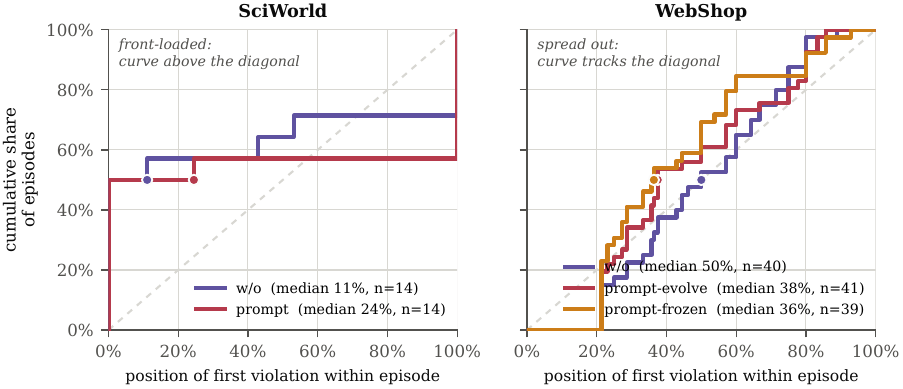}
\caption{Cumulative distribution of the first violation's relative position within an
episode. Curves above the diagonal indicate violations concentrated earlier in the episode.}
\label{fig:timing}
\end{figure}

\paragraph{Primary metric.}
Per-action violation rates are affected by survivorship in SciWorld. The unconstrained variant
often terminates within one or two steps after a focus error, producing 220 actions across
20 episodes, whereas constrained variants continue for 625 actions and therefore have more
opportunities to act. We use the episode-level violation rate as the primary SciWorld metric.
WebShop has a fixed 15-step cap and comparable action counts across variants (579--744), so
episode- and action-level metrics are both appropriate and lead to the same conclusion.

\paragraph{Evaluator scope.}
Applying the unrestricted SciWorld specification reports 78 violations in the constrained
variant, but these activations arise from \texttt{no\_op\_repeat} disagreements on repeated
\texttt{look around} actions. That rule family remains unchanged across the conditions and is
not the component under evaluation. Restricting the evaluator to the focus-family primitives
gives the relevant count of zero. This example motivates evaluating a modified sub-module
only over the rule family it replaces.

\subsection{Scope-Aware Evaluation of Composite Agents}
\label{app:metrics}

Metrics aggregated over all actions can include behavior controlled by agent components that
are unchanged across the compared conditions. We observe this issue in three forms.

\paragraph{Dominance by an unchanged primitive.}
In the SciWorld fidelity calculation, 85 of 98 true positives come from
\texttt{no\_op\_repeat}, which is active in both variants. An aggregate $F_1$ score therefore
mostly reflects agreement on a rule family that is not under evaluation.

\paragraph{Scope spillover.}
False positives can occur on non-focus actions outside the scope of the focus guard. Under
the aggregate metric, increasing synthesis evidence appears to reduce fidelity
($F_1=0.222$); under focus-isolated precision, the ordering reverses
($0.778$ versus $0.625$).

\paragraph{Violation-rate contamination.}
Using the unrestricted specification to evaluate Section~\ref{sec:violation} produces
78 apparent violations in the constrained SciWorld variant. All are
\texttt{no\_op\_repeat} disagreements on repeated \texttt{look around} actions, while the
focus-family count is zero.

These cases show that action-level evaluation of a composite agent should be restricted to
the scope of the sub-module being changed. Error types should also be interpreted according
to their costs, because a false block and a missed block can have substantially different
consequences in an irreversible action space.

\subsection{Selection Effects in Win-Rate-Based Validation}

The surviving lessons in the prompt-evolve variant can have high observed win rates
(e.g., \texttt{uses}=6 and \texttt{wins}=4), but the retention rule itself keeps only lessons
whose win rate exceeds 50\%. High win rates among retained lessons are therefore partly a
consequence of conditioning on survival. With a base success rate of 0.22, some retained
lessons may reflect favorable short sequences rather than reliable improvements. Evaluations
of win-rate-gated memories should therefore account for this selection effect.

\end{document}